\documentclass{article}

\usepackage[nonatbib, final]{ap_2018}

\usepackage[pdftex]{graphicx}
\usepackage[utf8]{inputenc} % allow utf-8 input
\usepackage[T1]{fontenc}    % use 8-bit T1 fonts
\usepackage{hyperref}       % hyperlinks
\usepackage{url}            % simple URL typesetting
\usepackage{booktabs}       % professional-quality tables
\usepackage{amsfonts}       % blackboard math symbols
\usepackage{amsmath}
\usepackage{nicefrac}       % compact symbols for 1/2, etc.
\usepackage{microtype}      % microtypography
\usepackage{xcolor}
\usepackage{multirow}

\title{Alquist 2.0: Alexa Prize Socialbot Based on Sub-Dialogue Models}

\author{
  Jan Pichl \\
  FEE, CTU Prague\\
  Prague, Czech Republic \\
  \texttt{pichljan@fel.cvut.cz} \\
   \And
   Petr Marek \\
   FEE, CTU Prague\\
   Prague, Czech Republic \\
   \texttt{marekp17@fel.cvut.cz} \\
   \And
   Jakub Konr\'ad \\
   FEE, CTU Prague\\
   Prague, Czech Republic \\
   \texttt{konrajak@fel.cvut.cz} \\
   \And
   Martin Matul\'ik \\
   FEE, CTU Prague\\
   Prague, Czech Republic \\
   \texttt{matulma4@fel.cvut.cz} \\
   \And
   Jan \v{S}ediv\'{y} \\
   CIIRC, CTU Prague \\
   Prague, Czech Republic \\
   \texttt{jan.sedivy@cvut.cz} \\
}

\begin{document}

\maketitle

\begin{abstract}
  This paper presents the second version of the dialogue system named Alquist competing in Amazon Alexa Prize 2018. We introduce a system leveraging ontology-based topic structure called topic nodes. Each of the nodes consists of several sub-dialogues, and each sub-dialogue has its own LSTM-based model for dialogue management. The sub-dialogues can be triggered according to the topic hierarchy or a user intent which allows the bot to create a unique experience during each session.
\end{abstract}

\section{Introduction}
In this paper, we present the second iteration of Alquist socialbot, a conversational system designed to converse coherently and engagingly with humans on popular topics. 

Our system focuses on coherent, informative and engaging dialogues centered around the large scale of topics including pop culture, sports, technology, etc. We have designed the system to be highly modular and the user experience to be as non-repetitive as possible. We achieve this by engaging in dialogues that are highly interconnected.  The bot switches between topics seamlessly and maintains awareness of the ever-changing context. We considerably improved the system's NLU module, mainly focusing on intent and entity recognition, while employing state of the art methods like dialogue acts \cite{pichl2018dialogueacts}. 

The Alquist socialbot builds on the experience and knowledge gained from Amazon Alexa Prize 2017 \cite{pichl2018alquist}. However, it is a brand new system created from scratch for Alexa Prize 2018. The results of Alexa Prize 2017 \cite{ram2018conversational} showed that meaningful conversation with a socialbot is possible, however, there is still a long way to go for artificial intelligence to handle a complex human-like conversation.

In the paper, we first describe the high-level design of the system (Section \ref{system_overview}), then we describe each of the system components in detail (Section \ref{system_components}). We dedicate a section to the dialogue flow and dialogue design process (Section \ref{dialogue_flow}) as it is integral to the success of the socialbot. Finally, we describe the experiments (Section \ref{experiments}) we performed during the creation of the system and finally summarize our findings (section \ref{conclusion}).

\subsection{Innovations}

We propose several improvements in comparison to our bot competing in the Amazon Alexa Prize 2017. The current version of the bot works with small dialogue structures which can be easily grouped into nodes that we call ``topic nodes''. These nodes group the dialogues about the same topic as opposed to the last year's version, where there was one fixed tree structure for each topic node. Current version allows the bot to choose a different path through the topic node for each session. Moreover, the user is able to start a specific sub-dialogue of the given topic node.

Another innovation is the improved workflow for adding new content. We implemented a custom web-based editor where the dialogue structure can be created and changed easily. Based on the structure, the dialogue manager (DM) model is trained, and auxiliary Java code is generated. This process allows us to divide the dialogue design part and the implementation part (which is not always necessary).

We present our DM as a novel approach to drive the open-domain dialogue. We have a separate DM model for each sub-dialogue. This model is inspired by Hybrid code networks \cite{williams2017hybrid} which were originally designed for task-oriented dialogues. Selecting the particular sub-dialogue (and corresponding DM model) is handled by a topic switch detector and intent detector. The topic nodes are formed in a tree structure which allows the bot to select similar topics when there is no strict user initiative. Each of the models used during the dialogue management uses contextual information.

\section{System overview}
\label{system_overview}

\begin{figure*}[ht]
\begin{center}
\includegraphics[width=\linewidth]{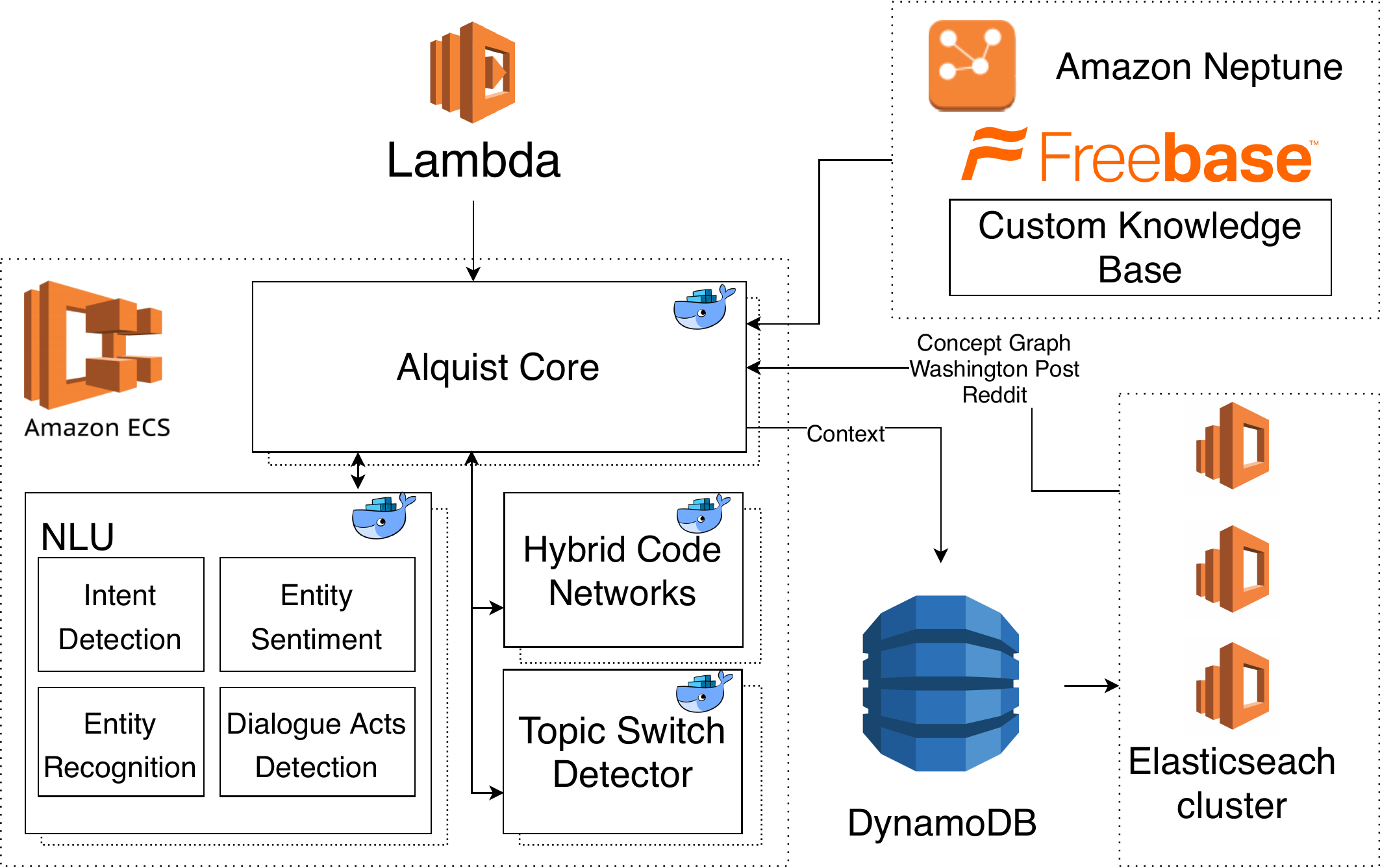}
\caption{The system architecture schema} 
\label{fig:architecture}
\end{center}
\end{figure*}

The system consists of several components which are visualized in Figure \ref{fig:architecture}. Each component is described in detail in Section \ref{system_components}. The system uses AWS lambda function which is connected to Alexa Skill. The lambda function only sends the requests as is to the Alquist Core component.

The core system is a Java application which is connected to each of the remaining components. NLU module encapsulates several models such as intent detection, entity recognition, entity sentiment, and dialogue acts. The results of each model can be retrieved by a single request. Hybrid code network is a module which contains DM models for all sub-dialogues. Topic switch detector is a model which decides whether the user still wants to talk about a current topic or not. All of these four modules are standalone docker containers. This design takes advantage of AWS Code Pipeline\footnote{https://aws.amazon.com/codepipeline} based continuous integration. The usage of AWS Code Pipeline was inspired by the Cobot which is a socialbot framework provided for each team competing in the Alexa Prize 2018 by Amazon. The dotted shadow rectangles behind each docker container showed in Figure \ref{fig:architecture} illustrate the auto-scaling ability of each component. By default, each component is launched as a single container. Based on the CPU and memory usage the additional containers are subsequently launched.

All the information (context) gathered during each dialogue turn is persistently saved in the DynamoDB. The information includes all of the annotations, message, generated response, current topic, current state, etc. For a better analysis and faster search, the content of the DynamoDB is duplicated in a Elasticsearch index. The Elasticsearch also contains crawled data such as Washington Post articles, Reddit content and Microsoft Concept Graph \cite{concept1}.

That last component is an RDF knowledge graph stored in Amazon Neptune. It contains the last dump of Freebase \cite{Bollacker:2008:FCC:1376616.1376746} enriched by our custom data.

\subsection{Information flow}

The following list describes when the individual components are triggered during the processing of a single dialogue turn.

\begin{enumerate}
    \item The request from Alexa Skill is sent to the \textbf{Lambda function}. The request data are passed to the Alquist Core without any modifications.
    \item The \textbf{Alquist Core} receives the request and based on the session ID, it loads the context (up to 20 previous turns) from DynamoDB. If there is no previous turn (the user just started the conversation), a welcome dialogue is triggered with an empty context history.
    \item The user message is annotated by all of the models in the \textbf{NLU module} and the Topic Switch Detector is triggered.
    \item If the \textbf{Topic Switch} model detects topic switch, a new sub-dialogue is selected based on the topic nodes assigned to the current entity and intent combination.
    \item If the \textbf{Topic Switch} model does not detect topic switch, the bot continues with the previous sub-dialogue, or it suggests a new one if the previous one is finished.
    \item The user message is sent to the \textbf{Dialogue manager} which produces response according to selected sub-dialogue. 
    \item Before the pipeline finishes, the bot stores the context in the DynamoDB.
\end{enumerate}

\section{System components}
\label{system_components}

\subsection{Context}

Context is an object holding all relevant information for the current dialogue turn. At the beginning of each turn, user utterance, user ID and session ID are saved into the object. Each object contains references of up to 20 previous context objects (retrieved according to the session ID). Note that all the stored information (NLU annotation etc.) is dialogue turn exclusive. However, context object contains a nested object called Session Attributes which is propagated from the current turn to the next one. At the end of each turn, the context object is persistently stored in DynamoDB including all nested classes such as Session Attributes.

There is a separate object called User Attributes which can store relevant information for a given user ID. This information can be used across all sessions. This object is stored in a separate table in DynamoDB.

\subsection{Knowledge base}

To obtain content for the dialogues, we scrape data from various sources, for example, Washington Post or Reddit. Washington Post articles are downloaded daily through an API provided by Amazon. Reddit posts from the subreddit ``Today I learned'' are used as trivia about people and other entities. We designed our knowledge base in order to add structure to this data. We chose the RDF-triples format to expand it with information already converted to it, namely a dump of Freebase.
Data integration between Freebase and our data (articles from now on) is done as follows: First, each article is searched for named entities. Articles from certain sources already contain this information, but the rest needs to be processed with a named entity recognition tool. We use the NLTK library for this because of its speed and certain features. Once the entities are obtained, they are linked to their Freebase IDs using a query to an implementation of a Freebase fuzzy label lookup\footnote{https://github.com/brmson/label-lookup}. Finally, the articles and Freebase entities are linked using a simple ontology of our own design.
The knowledge base runs on Amazon Neptune and is periodically updated with new articles. It can be queried with SPARQL query language. One improvement over Elasticsearch, where we stored the data initially, is the fact that when we look for an article about an entity, we can be sure that it is mentioned there. 

\subsection{NLU module}

NLU module is a single endpoint containing several models such as entity recognition model, intent detection model, dialogue act detection model, and sentiment analysis model. The results of all included annotations can be retrieved using a single request, alternatively each annotation result can be requested individually. Since most of the models take a user utterance as input (except for entity sentiment analysis), it is better to request all the annotations at once to avoid the network overhead.

\subsubsection{Entity recognition}
\label{sub:entity}

Entity recognition is a task which gives a label for every single word from a given utterance. We use a \textit{inside, outside, begin} (IOB) \cite{ramshaw1999text} schema commonly used in Named Entity Recognition (NER) task with an addition of type of entity (e.g. \textit{B-movie}). The only difference to the traditional NER task is that we do not require the entity to be a strict named entity which can be mapped, e.g., to a Wikipedia article. For example, in the sentence \textit{I want to talk to you about my life}, the word sequence \textit{``my life''} is marked as an entity although it is not a proper named entity. This approach allows the bot to start a dialogue about abstract topics such as \textit{``my life''}. 

Additionally, we want to recognize a correct entity type from the utterances where it is possible. This type-inference is only based on the current utterance, thus it should be possible for the model to recognize it. For example, in the sentence: \textit{I want to talk about Matrix}, our model should label \textit{``Matrix''} as a Generic Entity since it is not possible to know that Matrix is a movie without external knowledge. On the other hand, in the sentence \textit{Let's chat about the Matrix movie}, it should label \textit{``Matrix''} as a movie.

For the sequence tagging task, we use BI-LSTM-CRF \cite{huang2015bidirectional} model. The input is a single user utterance where each word is represented by a word embedding. The embeddings are followed by bidirectional LSTM layer which is then connected to a fully connected layer. The sentence features extracted by the previous layers are then fed into the CRF layer. The number of predicted classes is two times the number of types (one for B and one for I) plus one (for O). As training data, we use a manually labeled dataset of utterances gathered during conversations with the real users. We use our own annotation tool to make the process as fast as possible.

\subsubsection{Intent detection}
\label{sub:intent}

Intent detection classifies each utterance into one of the predefined classes. These classes are related to the sub-dialogues which the bot is capable of talking about. The classes are, for example: ``tell topic'', ``change name'', ``tell news'', etc. Detected intent combined with recognized entities is used in the decision which sub-dialogue should be triggered. There is not a strict boundary between entity and intent during the dialogue design. For example, the sentence \textit{Let's talk about rock music} can be labeled with \textit{tell\_topic} intent and \textit{rock music} entity or the intent can be \textit{tell\_about\_music} and entity can be \textit{rock}. Those two approaches are equal as long as we stay consistent. Note that a new sub-dialogue is triggered based on the entity and intent only if the topic switch is detected (based on the contextual data) as described in subsection \ref{topic_graph}.

We use multi-channel convolutional neural network \cite{kim2014convolutional} as a model for the intent detection. The words from input utterances are represented as the word embeddings. Embeddings are followed by five channels of convolutions, followed by a max pooling layer. The last two layers are fully connected layers. The output of the model is the probability of each intent class. The model is trained on a dataset which is a combination of the utterances from real conversations and utterances generated from templates.

As we mentioned earlier, entity recognition and intent detection are highly related. We run a few experiments including a combined model for both of the tasks. These experiments are described in section~\ref{experiments}.

\subsubsection{Dialogue act detection}

Dialogue act detection is also the utterance classification task. Unlike the intent detection, the dialogue act classes are not related to the specific dialogues, but they can be used across various NLP tasks. The classes describe whether the utterance is a statement, question, acknowledgment, etc. We use commonly used Switchboard dataset \cite{godfrey1992switchboard} which is a phone call transcription annotated by the dialogue acts. Original dataset contains over 200 classes which are clustered into 43 classes based on the predefined rules\footnote{\url{https://web.stanford.edu/~jurafsky/ws97/manual.august1.html}}. We trained our model using these 43 classes.

We use the same model architecture as for the intent detection. We do not use predicted dialogue acts directly during the conversations, but we rather use a feature vector which is an output of the second-to-last fully connected layer. This feature vector is input to the dialogue manager as described in the subsection \ref{dialogue_manager}. However, as reported in \cite{pichl2018dialogueacts}, there is an improvement in dialogue act detection accuracy when the contextual information is incorporated. We did not include the contextual information since the model of dialogue manager takes a window of the utterances and dialogue acts as the input, and it is the only usage of the dialogue acts in the system.

\subsubsection{Entity sentiment}
Entity sentiment module is tasked with giving the socialbot context for a recognized entity and forming the socialbot's opinion on the given entity. For each detected entity we search twitter via TwitterAPI for recent tweets containing mentioned entity. We perform sentiment analysis on gathered tweets and compute a mean value from received results. We use this value as a basis for the socialbot's initial opinion on the given entity.

For sentiment detection, we first clean the tweet of non-word strings and tokenize it, then convert the individual tokens to their embedding representation. We use them as an input to a bidirectional recurrent (GRU) neural network layer. Our model utilizes a dense layer from which we obtain estimated sentiment. The model then determines the sentiment of the tweet on the scale from 0 (negative) to 1 (positive). We have trained our model on two separate datasets the IMDB sentiment dataset \cite{maas-EtAl:2011:imdb} and the twitter sentiment classification dataset \cite{go2009twitter}. In the Section \ref{experiments} we compare the models trained on the different datasets. We are currently using the model trained on the IMDB movie review dataset.

\subsection{Topic graph}
\label{topic_graph}
\begin{figure*}[ht]
\begin{center}
\includegraphics[width=\linewidth]{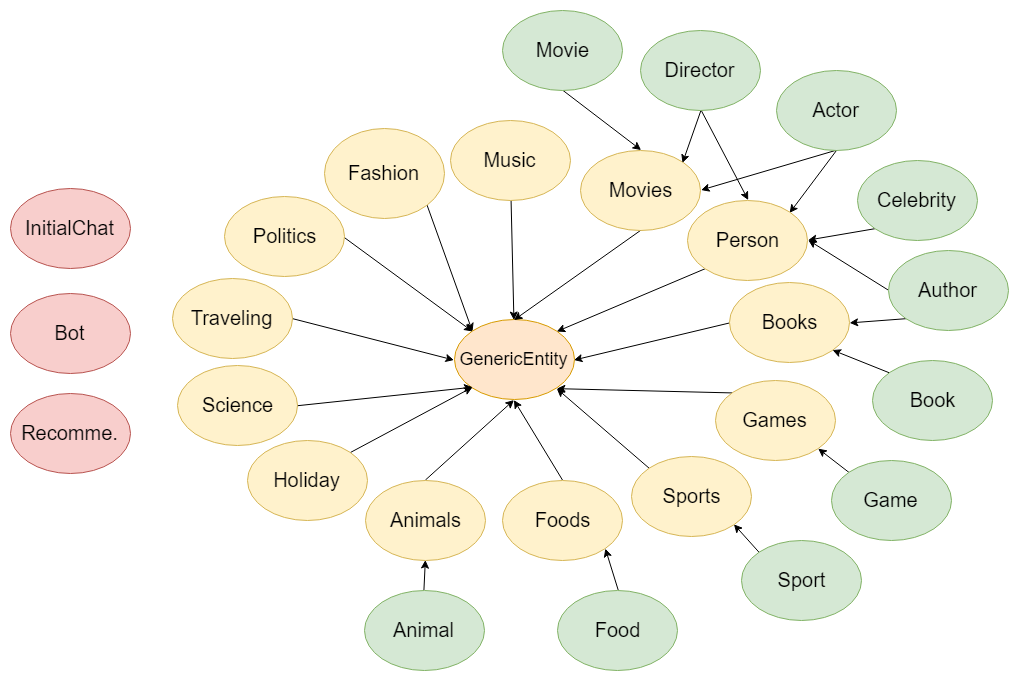}
\caption{Schema of Topic graph. Yellow nodes contain dialogues about a topic. Green nodes contain dialogues about an entity of its type. Orange node GenericEntity contains dialogues about entities with unknown type. Red nodes are special topic nodes containing dialogues about the bot, user, and Initial chat. These nodes are not connected to the rest of the Topic graph.} 
\label{fig:topic_graph}
\end{center}
\end{figure*}
The Topic graph contains a graph structure of topics, sub-dialogues, and their interconnections. We show the structure of topics in our Topic graph in Figure \ref{fig:topic_graph}. It consists of topic nodes and sub-dialogues. Each topic node has assigned one or more sub-dialogues. For example, ``Movies'' topic node has assigned sub-dialogues about movies in general, like \textit{``Where do you watch movies''}, \textit{``What is your the most favorite movie''} or \textit{``Are you a big movie fan''}. The Movie topic node has assigned sub-dialogues about some specific movie like \textit{``Actor starring in the movie''}, \textit{``Director of the movie''} or \textit{``What is your favorite part of the movie''}. Each of these sub-dialogues is implemented as a model in Hybrid code networks.

Topic nodes are connected by oriented edges, which point from more specific to less specific topic nodes. There is, e.g., an edge from Movie to Movies, or from Director to Movies and Person. If we detect that the user wants to talk about Steven Spielberg, and we know that he is a director, we can select any sub-dialogue from Director node and all the nodes which can be reached from the Director node (i.e., Person, Movies). The probability of node selection is based on the distance from the (current) Director node. A shorter distance means a higher probability. If the selected sub-dialogue ends, the topic graph automatically selects a new sub-dialogue from the Director node or the connected nodes. This method of active selection of new sub-dialogues keeps the user engaged. We have a special node we call ``GenericEntity'' for all entities of unknown type. It is the least specific topic node. It means that there is no oriented edge from GenericEntity node to any other node. It contains three sub-dialogues: \textit{``Funfact''}, \textit{``Shower-thought''} and \textit{``News''}. These sub-dialogues do not require any specific knowledge about the entity. They select the content of sub-dialogues by a text search of the entity name. This method allows us to maintain a conversation about any existing entity.

There is also a Recommendation topic node. Its purpose is to suggest new topics of the conversation to the users. This node is used only if users themselves do not specify the topic of the conversation, or we run out of sub-dialogues for the topic which they requested.

\subsection{Dialogue manager}
\label{dialogue_manager}

\begin{figure*}[ht]
\begin{center}
\includegraphics[width=\linewidth]{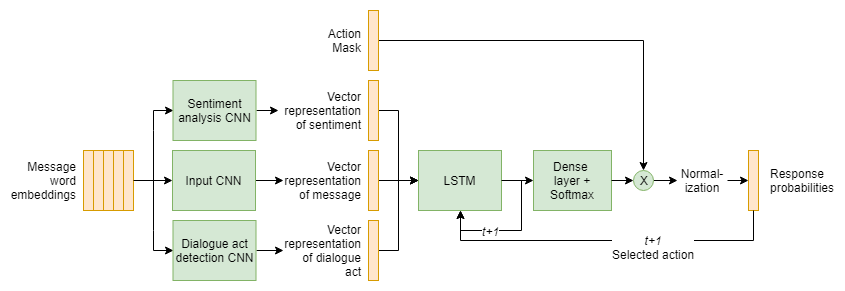}
\caption{Schema of our implementation of Hybrid code networks} 
\label{fig:hybrid_code_networks}
\end{center}
\end{figure*}
We use modified Hybrid code networks (HCN) described in \cite{williams2017hybrid} as our dialogue manager. HCN are a dialogue manager which combine an RNN with domain-specific knowledge encoded as software and system action templates. Model's task is to select the best response based on the input message and the context of the dialogue. HCN require less training examples compared to other end-to-end approaches thanks to domain-specific code but also retains the benefits of end-to-end learning. The model obtained state-of-the-art performance on the bAbI Dialogue tasks \cite{bordes2016learning}. These three properties (ranking of handmade responses, low data requirements, and end-to-end learning) are the main reasons why we decided to use HCN as our dialogue manager.

Our modified implementation of Hybrid Code networks consists of following components: input convolutional neural networks, recurrent neural network, and domain-specific code implementing text actions, functions, action masks and \textit{can start} methods.

The model obtains the input message which is featurized in three ways. Firstly, we use input convolutional neural network described in \cite{kim2014convolutional} with pretrained fastText embeddings \cite{bojanowski2016enriching}. We obtain the featurized output from second last layer of CNN. The CNN is trained on the training data of the sub-dialogue. Secondly, we use the architecture of \cite{kim2014convolutional} again, but the weights are pre-trained on sentiment analysis task. The pre-trained weights are frozen during training on the sub-dialogue. We again obtain the featurization of input message from the second last layer of CNN. Thirdly, we use the pre-trained model for dialogue act detection, from which we extract the output of second last layer. The weights of the model are frozen during training. 

The featurization variants of input message are concatenated into a single vector. We furthermore concatenate  response class predicted in the previous step. The resulting vector is fed into RNN. The timesteps of RNN are unusual. Instead of going across the values of the single input vector, the timesteps go across the input messages. This allows the model to learn the representation of the dialogue's context. Our RNN layer consists of LSTM \cite{hochreiter1997long} cells.

The vector output of RNN is element-wise multiplied by action mask vector. The action mask vector consists of zeros and ones. Its purpose is to prohibit some actions by assigning them a zero probability. The action mask vector is produced by action mask code. It consists of a set of rules. They prohibit usage of responses which don't directly follow the last used response in the dialogue. The element-wise multiplied vector is fed to softmax layer which computes probabilities of responses. We select the response with the highest probability.

There are two types of responses, text responses and functions. Text responses are directly returned as the response, no more processing is required. The functions are represented as some code which needs to be executed. The result of the code must be the class of the following response, which is a text response or another function. The function code can meanwhile arbitrarily modify values saved in the context.

The text response may contain text actions. Such response can be \textit{``Movie was directed by \{say\_director\}''}, where the text action is \textit{\{say\_director\}}. The text actions are replaced by text actions code before they are presented to the user. The text action code can also modify values saved in the context.

The last part of our HCN implementation is \textit{can start} code. It determines whether the dialogue can be started based on the values saved in the context. For example, if the dialogue requires the director's name, and the name is unknown (e.g., not found in DB), the \textit{can start} code flags the dialogue as not able to start. Another reason for this flag can be that the sub-dialogue has been executed previously and we do not want to repeat it. In such case, the topic graph has to select different dialogue to execute.

\subsubsection{HCN training}
We found the best set of hyperparameters for the HCN model on the validation set of bAbI Dialogue Task 6 by Bayesian hyperparameter optimization prior to the training. The best parameters for several architectures are presented in Table \ref{hyperparameters}. We generate a dataset of all possible transitions through the dialogue, which was created in our graphical editor. We split the dataset into training, validation and testing examples. One example is equivalent to one whole transition through the dialogue. Training part contains 80\% of examples, and validation and testing both 10\% of examples. We use three-fold cross-validation on training examples to find the best number of training epochs. We limit the maximum number of epochs to 12. We train the model on whole training dataset for the number of epochs determined by the cross-validation. 

\subsection{Topic Switch detector}
\begin{figure*}[ht]
\begin{center}
\includegraphics[width=\linewidth]{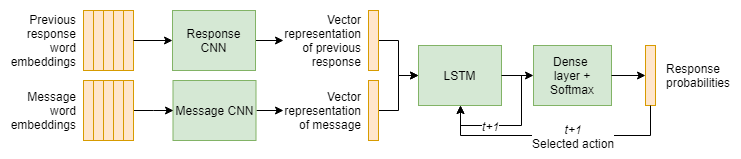}
\caption{Schema of Topic Switch detector} 
\label{fig:topic_switch}
\end{center}
\end{figure*}
Topic Switch detector is a component which determines whether the user wants to switch the topic of the conversation. If such request is recognized, we switch the topic of the conversation according to the detected intent and recognized entity. We use this model on top of the intent detection because a decision to switch the topic must be conditioned on the context of the conversation. Intent detection works only with the recent message. It is not sufficient, because message \textit{``I like pop music.''} can have different topic switch labels in different contexts. If socialbot asks user \textit{``What do you like?''}, we want to switch the topic to sub-dialogues about music. However, if we are in the middle of the dialogue about favorite music genre and the socialbot asks user \textit{``Which music genre is your favorite?''}, we do not want to switch topic because it leads to a restart of the conversation about music. But intent detector returns intent \textit{``Music''} for message \textit{``I like pop music.''} in both contexts. This is the reason why we use Topic Switch detector which is trained to make decisions not only on recent message but also on the context of the conversation.

The Topic Switch detector's model uses an architecture similar to the HCN model. It consists of two input CNNs \cite{kim2014convolutional} and an RNN. Model's inputs are the last dialogue turn's response and the current user's message. The input CNNs create vector representations of both inputs which are concatenated and passed to LSTM and softmax layer. Two output classes of the softmax layer correspond to the probability of user wanting to switch the topic and not wanting to switch the topic of dialogue.

The topic switch is trained on sub-dialogues from the Topic graph. We generate artificial conversations from training data for individual dialogues, in which we mix training examples of intent detection. We mark the turns into which we mix the intent example by class one which indicates the topic switch and the rest of turns by class zero. The model learns to predict these labels.

\section{Work-flow of dialogue creation}
\label{dialogue_flow}

One of the main features which we optimize for Alexa Prize 2018 was the complexity of adding new dialogue to the socialbot. The complexity of adding a new dialogue was fairly high for last year's Alquist due to the fact that dialogues were represented as state graphs. Not only the content of the dialogues but also the inner decision logic had to be hand-made. This added a significant amount of work, which limited the number of topics we were able to include in time. We were able to include 27 topics to this year's Alquist as opposed to only 17 topics in case of the previous year. This number of topics was achieved by a smaller team of four people instead of five and also in shorter time as Alexa Prize 2017 started in November 2016 and Alexa Prize 2018 started in February 2018. This was accomplished thanks to two major innovations, which are the Hybrid code networks and graphical dialogue editor.
Hybrid code networks significantly reduced the amount of work required to create inner logic. The graphical dialogue editor simplifies the development of dialogue's content.

In the next sections, we will go through the process of adding sub-dialogue about writer's popularity. This sub-dialogue can be executed if a user wants to talk about any writer. Such dialogue will be one of many sub-dialogues a writer.

\subsection{Adding topic node to Topic graph}
The first step is to add a topic node \textit{Writer} to the Topic graph. This will enable us to add multiple sub-dialogues about any writer and transition between them. We represent each topic node as a YAML file, which contains the list of sub-dialogues and node's parents. We add \textit{writer\_popularity} sub-dialogue to the list of sub-dialogues. We add topic nodes \textit{Person} and \textit{Books} to its parents. Connection to \textit{Person} and \textit{Books} nodes allows us to smoothly transition to sub-dialogues assigned to these nodes, which contains dialogues of related topics to \textit{Writer}. 

\begin{figure*}[ht]
\begin{center}
\includegraphics[width=\linewidth]{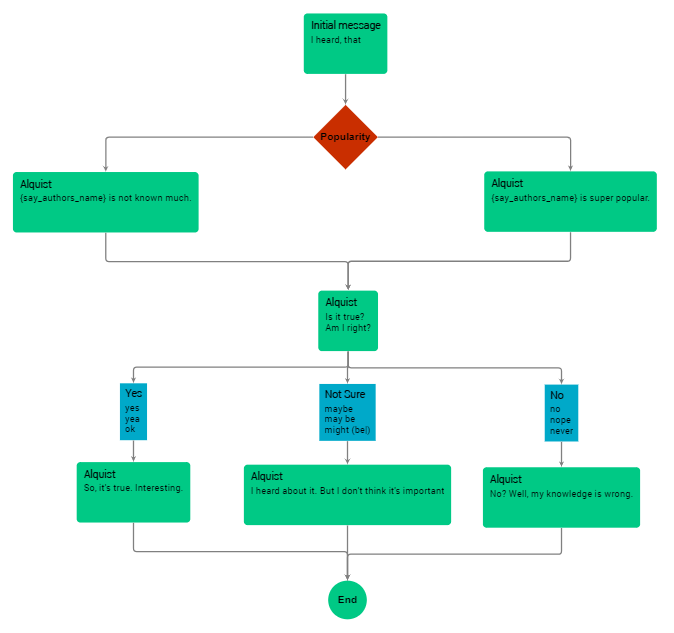}
\caption{The sub-dialogue about writer's popularity implemented in graphical dialogue editor. Green nodes are \textit{Bot} nodes, blue nodes are \textit{User} nodes and red nodes are \textit{Function} nodes.} 
\label{fig:graphical_dialogue_editor}
\end{center}
\end{figure*}

\subsection{Creating content of sub-dialogue in graphical dialogue editor}
The next step is to create structure and content of the dialogue in the graphical dialogue editor. We create it by placing and connecting \textit{Bot}, \textit{User} and \textit{Function} nodes. \textit{Bot} nodes represent the class of bot's responses. \textit{User} nodes represent the possible user's messages, which we use as the training examples for Hybrid code networks. Initially, we have to come up with the possible user's messages, but later we can enhance them by data from the real traffic. Both \textit{User} and \textit{Bot} nodes can contain multiple messages or responses. \textit{Function} nodes contain code that decides the next class of response. 

Finished dialogue is later converted to training examples for HCN. We do it by generating all possible transitions through the dialogue. The training script for HCN generates the files with model's weights, responses, and Java code templates of \textit{can start}, \textit{functions}, \textit{text actions} and \textit{action masks} methods. The file containing the model's weights and responses are uploaded to S3 bucket, and the rest of the files is moved to Java core. 

\subsection{Implementing code templates}
We have to implement Java code templates in the Java core. The first template is a \textit{can start} method. The purpose of this method is to signalize whether the dialogue can be executed. We prepare all the data which the dialogue needs in this method, and we check that the dialogue was not used before. In the case of our dialogue about writer's popularity, we check if we use this dialogue for a particular writer for the first time, and we make a request to retrieve the number of writer's fans from a database. The result of the request is saved to the context. If we use the dialogue for the first time and the request returns the count of fans successfully, the \textit{can start} method signalizes that the dialogue can be executed.

The code inside of \textit{function} templates decides which response class to use next based on the handwritten rules. It is an optional supplement to the decisions made by HCN. There are decisions in the dialogues, which depend on the external knowledge and not on the user's message. Such decisions are impossible to learn and must be implemented in code. We use the function to redirect the flow of the dialogue according to the number of writer's fans in our example. This can be implemented as a simple if-then-else rule comparing the value to some threshold.

String replacement is in the center of the \textit{text actions}. They are used to include information to responses, which can't be included at the time of dialogue creation. Such information is usually obtained from the database during runtime and inserted into the response. In our example the response can be \textit{``Writer has \{say\_fans\} fans.''}, where the text action is \textit{\{say\_fans\}}. The code implemented in \textit{text action} is responsible for replacing \textit{\{say\_fans\}} by actual value. This value is usually retrieved from the context, to which the code of \textit{can start} had saved it before the dialogue was executed.

The last code template is \textit{action mask}. This template prevents HCN from predicting non-logical response classes (those which are not connected in the designed sub-dialogue graph). \textit{Action mask} is completely generated by the training script of HCN. The code is fully customizable in theory, but we rarely do any changes to it in practice.

\subsection{Intent and entity annotation}

\begin{figure*}[ht]
\begin{center}
\includegraphics[width=\linewidth]{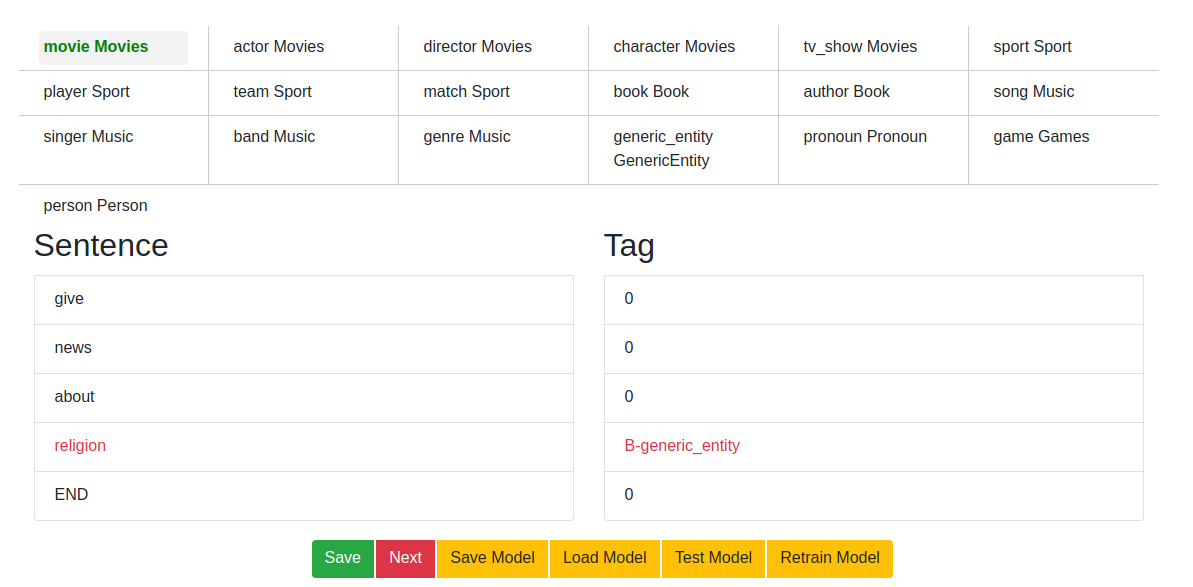}
\caption{UI of our entity labelling tool} 
\label{fig:entity_labelling}
\end{center}
\end{figure*}

To adapt the intent and entity models, we are constantly adding new samples into our training data based on the conversation logs. It is simple to annotate the intent since it is a single label for each utterance. On the other hand, for entity recognition, each word from the utterance needs to be labeled by the IOB plus entity type labels. To simplify this process, we created a web-based annotation tool. The user interface of the tool is shown in Figure \ref{fig:entity_labelling}. The tool allows us to select an entity type and the just click on the words to annotate them. When the model is trained, it can be used for the label suggestion during the actual annotation process. Then, we can accept whole labeling or correct mistakes, save the labeled data and retrain the model.

\subsection{Retraining Topic Switch detector}

The last step is necessary only if we added new intent. Otherwise, it is optional. The step is to retrain the model of Topic Switch detector to correctly detect the user's desire to switch the topic. The training script of Topic Switch detector downloads all dialogues from S3 bucket and takes training examples from the intent recognition. It combines these two datasets into training data. The trained model's weights are then uploaded to S3 bucket.

\subsection{Final remarks}
These are the only necessary steps which we need to take in order to add new dialogue. Most of the tasks are automatized or consists of creative work. The only technically challenging part is the coding of the java code templates. The work to add dialogue can be thus divided into a creative and a technical part. The creative part can be done by dialogue designer with no programming experience. The knowledge of programming is needed only for the technical part.

\section{Experiments}
\label{experiments}

\subsection{Intent and Entity}

We have experimented with several models for the intent and entity tasks. We trained separate models for each task and compared it with the combined model which has outputs for both intent and entity. We had 8,052 samples for the intent detection and 3,494 samples for the entity recognition at the time of writing this paper. We are continuously annotating new samples during the bot development. Most of the samples are gathered from dialogue logs. At the beginning of the development, a small portion of samples was generated from templates.

The models for entity and intent are described in Subsection \ref{sub:entity} and \ref{sub:intent} respectively. The combined model is a modification of the entity model. It has two stacked BI-LSTM layers. The last state of the first layer is the intent output and the sequence output of the second layer is fed into a dense layer which is followed by CRF.

We tried three different embeddings for each task, and each test was triggered ten times. We used GloVe \cite{pennington2014glove} with 50 and 300 dimensions a custom fastText embeddings with 100 dimension. The results are shown in Table \ref{intent_entity_experiments}. 

\begin{table}[ht]
  \caption{Testing results of intent, entity, and combined models. There are three models for each task which shares the same architecture, but they use different embeddings. The accuracy and sentence error rate fields contain mean value and standard deviation across 10 measurements. The results for the combined task contains values for the intent task/values for the entity task. The sentence error rate is the metric applied only for the entity recognition and shows how many of the sentences have at least one misclassified token.}
  \label{intent_entity_experiments}
  \centering
  \begin{tabular}{l|l|c|c|c}
    \toprule

     & Model     & Accuracy   & Sentence error rate & Training time \\
    \midrule
    \multirow{3}{*}{Intent} & GloVe50   &  $ 91.6\% \pm 0.9 $ & - & \textbf{37 sec}     \\
                            & GloVe300      & $\boldsymbol{94.8\% \pm 0.4} $ & - & 2 min 10 sec \\
                            & FastText      & $ 94.7\% \pm 0.4 $ & - & 47 sec    \\
                            \hline
    \multirow{3}{*}{Entity} & GloVe50   & $ 98.6\% \pm 0.1 $  & $ 17.0\% \pm 1.7 $  & \textbf{1 min 14 sec} \\
                            & GloVe300      & $ 98.7\% \pm 0.2 $ & $ 17.6\% \pm 1.7 $ & 1 min 53 sec   \\
                            & FastText      & $ \boldsymbol{98.8\% \pm 0.2} $ & $ \boldsymbol{14.9\% \pm 1.8} $ & 1 min 18 sec   \\
                            \hline
    \multirow{3}{*}{Combined} & GloVe50   & $ 93.0\% \pm 0.5 $ / $ 98.2\% \pm 0.2 $ &  - / $ 21.1\% \pm 2.1 $ & \textbf{3 min 28 sec}   \\
                              & GloVe300      & $ \boldsymbol{95.0\% \pm 0.4} $ / $ 98.3\% \pm 0.2 $ & - / $ 21.3\% \pm 2.5 $ & 5 min 7 sec  \\
                              & FastText      &  $ 93.5\% \pm 0.4 $ / $ \boldsymbol{98.9\% \pm 0.2} $ & - / $ \boldsymbol{14.2\% \pm 2.4} $ & 3 min 42 sec  \\
    \bottomrule
  \end{tabular}
\end{table}

It is not a big surprise that the models with smaller embeddings are trained faster. On the other hand, there is no such a big difference in training duration between GloVe300 and fastText. GloVe300 vectors are slightly better for intent detection whereas Fasttext seems to be better for entity recognition.

\subsection{Dialogue manager}

We evaluate three architectures of dialogue manager which are inspired by Hybrid code networks. Moreover, we compared the impact of using word2vec and fastText embeddings for each architecture. We evaluate the turn accuracy and dialogue accuracy of the models on bAbI Task 6 dataset and Alquist conversational dataset.

\begin{figure*}[ht]
\begin{center}
\includegraphics[width=0.7\linewidth]{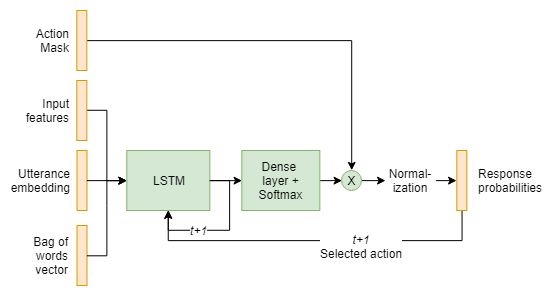}
\caption{Schema of Hybrid code network} 
\label{fig:SchemaOfHCN}
\end{center}
\end{figure*}

\begin{figure*}[ht]
\begin{center}
\includegraphics[width=\linewidth]{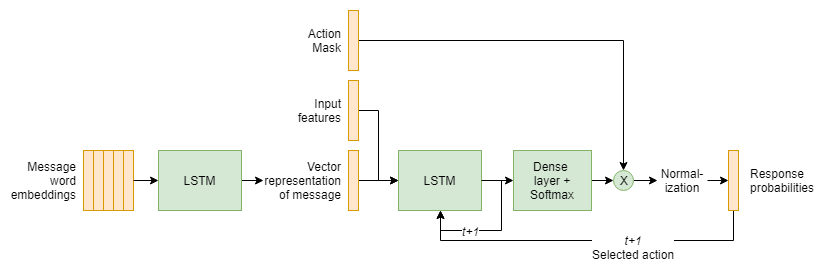}
\caption{Schema of Hybrid code network with recurrent input layer} 
\label{fig:SchemaOfHCNLSTM}
\end{center}
\end{figure*}

\begin{figure*}[ht]
\begin{center}
\includegraphics[width=\linewidth]{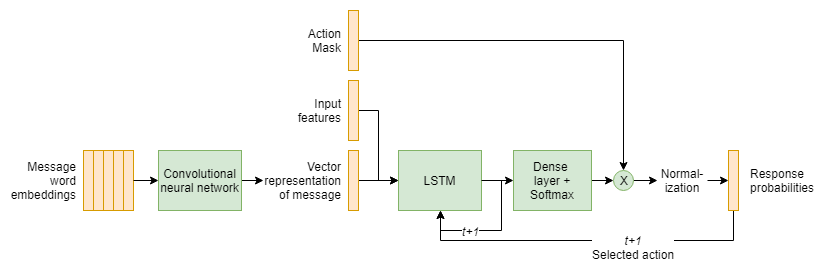}
\caption{Schema of Hybrid code network with convolutional input layer} 
\label{fig:SchemaOfHCNCNN}
\end{center}
\end{figure*}

\begin{table}[]
\caption{Set of best hyperparameters for each model founded by Bayesian hyperparameter optimization on the validation set of bAbI Dialogue Task 6 and achieved Turn accuracy}
\label{hyperparameters}
\begin{tabular}{c|cccccc}
\hline
\multicolumn{1}{l|}{}     & \multicolumn{6}{c}{\textbf{Model}}                                                                                                                                                                                                                         \\
\textbf{Hyperparameter}   & Word2vec & \begin{tabular}[c]{@{}c@{}}Word2vec\\ +CNN\end{tabular} & \begin{tabular}[c]{@{}c@{}}Word2vec\\ +RNN\end{tabular} & fastText & \begin{tabular}[c]{@{}c@{}}fastText\\ +CNN\end{tabular} & \begin{tabular}[c]{@{}c@{}}fastText\\ +RNN\end{tabular} \\ \hline
LSTM size                 & 85       & 109                                                     & 219                                                     & 55       & 245                                                     & 505                                                     \\
Convolutional filters     & -        & 6                                                       & -                                                       & -        & 21                                                      & -                                                       \\
LSTM dropout              & 0.92     & 0.79                                                    & 0.74                                                    & 0.85     & 0.80                                                    & 0.94                                                    \\
Input LSTM dropout        & -        & -                                                       & 0.91                                                    & -        & -                                                       & 0.97                                                    \\
Convolutional dropout     & -        & 0.84                                                    & -                                                       & -        & 0.72                                                    & -                                                       \\
Fully connected dropout   & 0.59     & 0.93                                                    & 0.98                                                    & 0.82     & 0.79                                                    & 0.76                                                    \\
Learning rate             & 0.001    & 0.005                                                   & 0.00005                                                 & 0.008    & 0.0001                                                  & 0.0003                                                  \\
Activation function       & tanh     & tanh                                                    & relu                                                    & relu     & relu                                                    & relu                                                    \\
Input activation function & -        & -                                                       & tanh                                                    & -        & -                                                       & tanh                                                    \\
Adam epsilon              & 1E-8     & 0.1                                                     & 1E-8                                                    & 1E-8     & 1E-8                                                    & 1E-8                                                    \\
Adam beta1                & 0.5      & 0.5                                                     & 0.9                                                     & 0.9      & 0.5                                                     & 0.5                                                     \\ \hline
Turn accuracy             & 71.3\%   & 70.4\%                                                  & 65.5\%                                                  & 69.4\%   & \textbf{71.5\%}                                                  & 68.0\%                                                  \\ \hline
\end{tabular}
\end{table}

\begin{table}[]
\centering
\caption{Testing accuracy of Hybrid code networks models}
\label{testingAccuracy}
\resizebox{\textwidth}{!}{%
\begin{tabular}{c|cc|cc}
\hline
                                 & \multicolumn{2}{c|}{\textbf{bAbI6}} & \multicolumn{2}{c}{\textbf{Alquist}} \\
\textbf{Model}                   & Turn Acc.         & Dialogue Acc.   & Turn Acc.       & Dialogue Acc.      \\ \hline
Bordes and Weston (2017) \cite{bordes2016learning}         & 41.1\%            & 0.0\%           & -               & -                  \\
Liu and Perez (2016) \cite{liu2017gated}             & 48.7\%            & 1.4\%           & -               & -                  \\
Eric and Manning (2017) \cite{eric2017copy}          & 48.0\%            & 1.5\%           & -               & -                  \\
Seo et al. (2016) \cite{seo2016query}               & 51.1\%            & -               & -               & -                  \\
Williams, Asadi and Zweig (2017) \cite{williams2017hybrid} & 55.6\%            & \textbf{1.9\%}  & -               & -                  \\ \hline
fastText                         & 57.6\%            & 0.8\%           & 86.9\%          & 51.7\%             \\
fastText+CNN                     & \textbf{58.9\%}   & 0.5\%           & 90.6\%          & 63.0\%             \\
fastText+RNN                     & 54.9\%            & 0.3\%           & 80.6\%          & 40.5\%             \\
word2vec                         & 57.4\%            & 0.4\%           & 92.2\%          & \textbf{68.0\%}    \\
word2vec+CNN                     & 56.3\%            & 0.1\%           & \textbf{92.6\%} & 67.8\%             \\
word2vec+RNN                     & 54.6\%            & 0.1\%           & 83.9\%          & 45.2\%             \\ \hline
\end{tabular}%
}
\end{table}

\subsubsection{Tested architectures of dialogue manager}
The first tested architecture of dialogue manager is the same as an architecture of HCN \cite{marek2018dialog}. It uses an average of word embeddings, bag-of-words vector and additional features as inputs. We concatenate these features and pass them to the LSTM layer. We element-wise multiply the output of LSTM layer by the vector of action mask. We pass this result to softmax function and select the response with the highest probability. The schema of architecture is in Figure \ref{fig:SchemaOfHCN}. The second architecture uses LSTM input layer instead of the average of word embeddings and bag-of-words vectors. It is visualized in in Figure \ref{fig:SchemaOfHCNLSTM}. The third architecture uses convolutional input layer inspired by \cite{kim2014convolutional}. The schema of this architecture is in Figure \ref{fig:SchemaOfHCNCNN}.

\subsubsection{Datasets}
The Dialog bAbI Task \cite{weston2015towards} Data is a dataset of conversations from restaurant reservation domain. It is used to test end-to-end dialog systems in a way that favors reproducibility and comparisons and is lightweight and easy to use. The dataset is divided into six tasks with increasing difficulty. We use task six because it is the only task which contains records of real-world conversations between humans and chatbot. It contains noisy and hard to learn dialogues due to voice recognition errors and non-deterministic human behavior. It is an ideal benchmark because we face the same challenges in Alexa Prize. The dataset contains 56 response classes, and it is split into 3,249 training dialogues, 403 validation dialogues, and 402 testing dialogues.

The Alquist conversational dataset is our private dataset collected from our previous version of socilabot competing in Alexa Prize 2017. The dataset consists of 37,805 dialogues between the user and the socialbot from the domain of books. There are 344,464 message-response pairs in total. The average length of dialogues is 9.11 pairs, the median is 7 pairs, and there are 23,633 unique responses. The dataset is also noisy and hard to learn because it contains voice recognition errors and part of the messages comes from uncooperative users. Messages from uncooperative users are hard to interpret or out of the domain of books. All of 23,633 responses can be clustered into 30 semantically unique responses. This reduction can be achieved thanks to the fact, that dialogues were represented as state graph. Each node in state graph correspondences to one of 30 semantically unique responses.

\subsubsection{Results}
We found the best set of hyperparameters for each architecture on the validation set of bAbI Dialogue Task 6 by  Bayesian hyperparameter optimization. They are presented in Table \ref{hyperparameters}. We trained the models with the best set of hyperparameters for 12 epochs on both datasets. The best model regarding turn accuracy on bAbI Task 6 dataset is model using convolutional input layer and fastText embedding vectors, which outperformed the baseline \cite{williams2017hybrid}. This model achieved turn accuracy of 58.9\%. The best model regarding turn accuracy on Alquist conversational dataset is model using convolutional input layer and word2vec embedding vectors, which achieved turn accuracy of 92.6\%. The complete results are presented in Table \ref{testingAccuracy}. 

\subsection{Sentiment}

We have trained our sentiment model separately on two different datasets the IMDB movie review dataset \cite{maas-EtAl:2011:imdb} and the sentiment140 dataset \cite{go2009twitter}. The IMDB dataset is a dataset containing long-form movie reviews with star ratings whereas the sentiment140 dataset contains tweets that have been annotated as positive or negative based on used emotes.

The model reached 0.90 accuracy on train set and 0.88 on the validation set for the IMDB dataset, and 
0.84 accuracy on train set and 0.83 accuracy on the test set for the sentiment140 dataset.

\begin{figure*}[ht]
\begin{center}
\includegraphics[width=\linewidth]{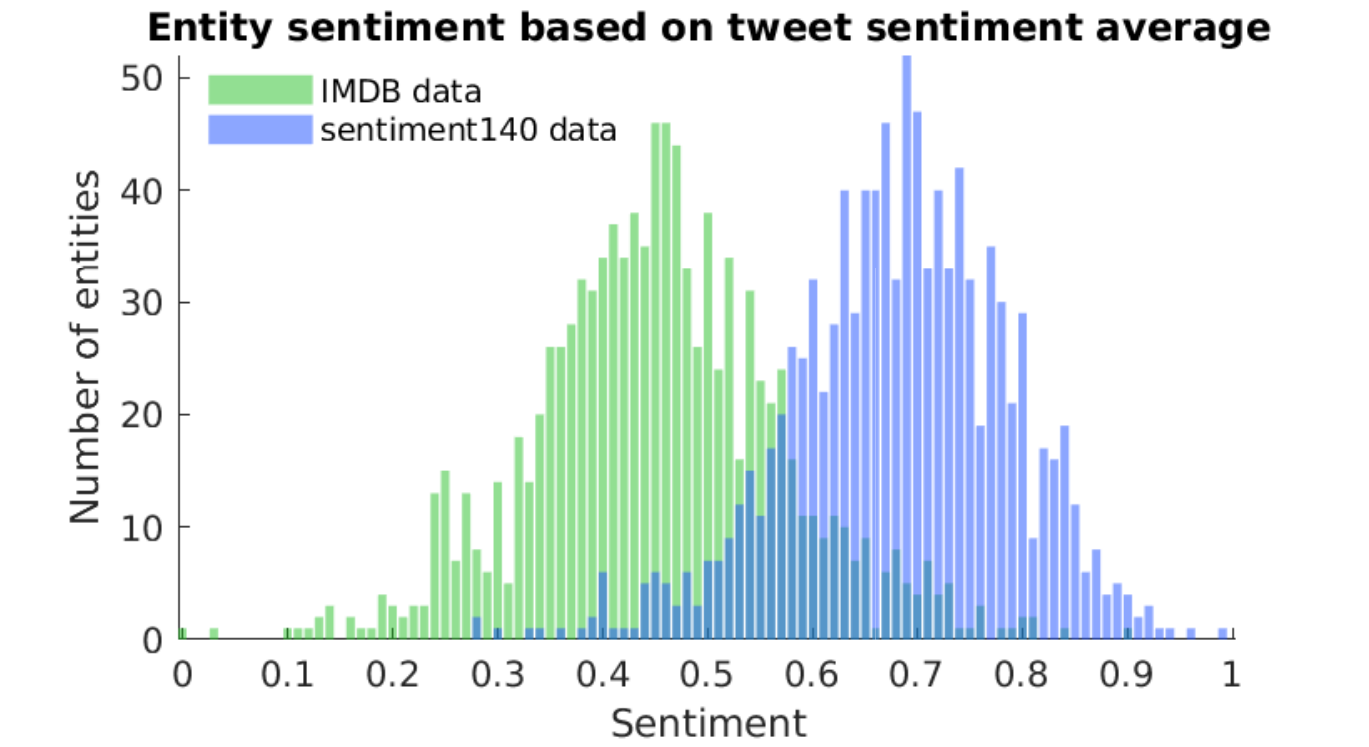}
\caption{The figure shows the histogram of detected entity sentiments for both trained models. Sentiment value 0 translates to the most negative and sentiment value 1 as the most positive.}
\label{fig:entity_sentiment}
\end{center}
\end{figure*}

We compared the models to see how do the detected sentiments differ for tweets containing entities recognized by our system. From the figure \ref{fig:entity_sentiment} it is clear that the model trained on sentiment140 data heavily skews towards positive sentiments for various entities. While we believe that could be helpful in order to keep interactions with user positive, the model often fails to recognize desired negative sentiments for serious and generally negative topics (see Table \ref{sentiment_table}). Due to this, we are currently using the model trained on the IMDB dataset.
\begin{table}[t]
  \caption{Sentiment values for entities with general negative connotations}
  \label{sentiment_table}
  \centering
  \begin{tabular}{l|c|c}
    \toprule

    Entity     & IMDB data     & sentiment140 data \\
    \midrule
    terrorism   & 0.38  & 0.53   \\
    Hitler      & 0.28 & 0.64    \\
    murder      & 0.24 & 0.38    \\
    \bottomrule
  \end{tabular}
\end{table}

% \subsection{Some nice table from data}

\subsection{AB testings}
We performed several AB testings during the competition, two of which brings interesting results. The first AB testing measured the impact of length of initial chat. The second measured impact of the active selection of dialogues.

\subsubsection{Initial chat}
The first AB testing compared two variants of initial chat which starts all conversations. The initial chat of A variant contained three dialogues: \textit{``What's your name?''}, \textit{``How are you?''} and \textit{``What are your hobbies?''}. We added five more dialogues in the B variant: \textit{``How is your family doing?''}, \textit{``Do you plan any vacation?''}, \textit{``How is it going at work?''}, \textit{``Do you have any pet?''} and \textit{``Have you tried other socialbots?''}. The B variant contained eight dialogues in total and was substantially longer. Our hypothesis was that longer initial chat would lead to longer conversations. The AB testing was active from July 1 to July 7. 

Both variants have the same weight of 50\%, and we collected the results from 906 dialogues for A variant and 883 dialogues for B variant. We included only dialogues for which users gave a rating. We present the results in Table \ref{initialChatAB}. Both variants achieved nearly identical results in all metrics. The initial hypothesis was not confirmed. We kept the variant A which contains only three dialogues.

\begin{table}[h]
\centering
\caption{AB testing of length of initial chat}
\label{initialChatAB}
\begin{tabular}{c|c|cc|c}
\hline
\textbf{}                   & \textbf{\begin{tabular}[c]{@{}c@{}}Avg Feedback \\ Rating\end{tabular}} & \multicolumn{2}{c|}{\textbf{\begin{tabular}[c]{@{}c@{}}Avg Duration \\ of Conversations\end{tabular}}} & \textbf{\begin{tabular}[c]{@{}c@{}}Avg Number \\ of Dialog Turns\end{tabular}} \\
\textbf{Variant}            &                                                                         & median                                        & 90th Percentile                                        &                                                                                \\ \hline
Three dialogues (variant A) & 3.03                                                                    & 2:54                                          & 13:18                                                  & 13.1                                                                           \\
Eight dialogues (variant B) & 3.01                                                                    & 2:55                                          & 13:18                                                  & 13.1                                                                           \\ \hline
\end{tabular}
\end{table}

\subsubsection{Paraphrasing}
The Paraphrasing AB test measures the effects of paraphrasing. Paraphrasing is a restatement of the user's message which we prepend to the response. The paraphrasing is implemented as a collection of rules which transforms the message into paraphrased form. These rules substitute ``you'' for ``I'', ``your'' for ``my'', ``if I am'' for ``are you'' or ``your parents'' for ``my mum and dad'' for example. These rules are applied only if the message contains word ``I'' or ``you'', its length is between two and nine words and with 50\% probability.

We test two variants. Variant A does not use paraphrasing while Variant B does. Both variants were applied to 50\% of users. We collected 3730 dialogues for variant A and 3681 for variant B. We include only dialogues for which users provided a rating. The results are presented in Table 6. The result shows that paraphrasing increases user rating.

\begin{table}[h]
\centering
\caption{Paraphrasing AB test}
\label{my-label}
\begin{tabular}{c|c}
\hline
\textbf{Variant}             & \textbf{\begin{tabular}[c]{@{}c@{}}Avg Feedback\\ Rating\end{tabular}} \\ \hline
Paraphrasing off (variant A) & 3.49                                                          \\
Paraphrasing on (variant B)  & 3.53                                                          \\ \hline
\end{tabular}
\end{table}

\subsection{Amount of trivia}
Amount of trivia AB test measures the effect of the number of fun facts, shower thoughts and news which we present to a user in a row. We test three variants. Variant A allows only single trivia dialogue in a row. Variant B allows two trivia dialogues following each other. Variant C allows three trivia dialogues in a row. We collected 1569 dialogues for variant A, 1607 dialogues for variant B and 1513 dialogues for variant C. We include only dialogues for which users provided a rating. The results are presented in Table 7. The result shows that a smaller number of trivia dialogues in a row increases user rating.

\begin{table}[h]
\centering
\caption{Amount of trivia AB test}
\label{my-label}
\begin{tabular}{c|c}
\hline
\textbf{Variant}            & \textbf{\begin{tabular}[c]{@{}c@{}}Avg Feedback\\ Rating\end{tabular}} \\ \hline
Trivia amount 1 (variant A) & 3.47                                                                   \\
Trivia amount 2 (variant B) & 3.44                                                                   \\
Trivia amount 3 (variant C) & 3.42                                                                   \\ \hline
\end{tabular}
\end{table}

\subsubsection{Entity switching}
The entity switching feature was intended to suggest a new entity to talk about in case the dialogues about the current entity were about to run out. This should prolong the conversation by introducing variety instead of piling further and further pieces of trivia on the user. The new entities were chosen from those included in already discussed news articles or trivia so as to be related to the previously used entities, i.e., if the bot read an article about Michael Jackson to the user, and that article mentioned David Bowie as well, the bot could suggest changing the topic of conversation from Jackson to Bowie. The entities related to the article were obtained beforehand, either having been already provided in case of Washington Post articles or by using Named Entity Recognition techniques on the pieces of trivia. The relations between articles and entities were stored in our RDF knowledge base then retrieved at runtime using a SPARQL query.

We experimented with two variants of this test. In the first variant, the suggestions were never made. In the other variant, the suggestion was made after reading the article although not every time an article was read, only in about 30\% of cases.  Both variants had an equal probability of 50\% of happening. This testing was active from October 12 to October 15. Variant A was seen by 3549 users and the variant B by 3645 users. The average rating for each variant is shown in Table \ref{entitySwitchAB}.

\begin{table}[h]
\centering
\caption{AB testing of entity switching}
\label{entitySwitchAB}
\begin{tabular}{c|c}
\hline
\textbf{}                   & \textbf{\begin{tabular}[c]{@{}c@{}}Avg Feedback \\ Rating\end{tabular}}   \\
\textbf{Variant}                                                                                    &                                        \\ \hline
Three dialogues (variant A) & 3.55                                                                                                                              \\
Eight dialogues (variant B) & 3.51                                                                                                                                                                                              \\ \hline
\end{tabular}
\end{table}

\section{Conclusion}
\label{conclusion}

We described the second version of the conversational bot Alquist. Even though the goal of the Alexa Prize competition 2018 is the same as in 2017, the system has been redesigned which helped us to develop more engaging conversations with less effort. Based on the experience gathered during the previous year, we have started to focus on the individual topics and the corresponding sub-dialogues. We proposed a novel approach to dialogue management by creating a separated model for each sub-dialogue. On top of that, the system is driven by recognized entities, intents and topic switch detection. Every model is created by generated data at the beginning, allowing us to fine-tune it later according to the real traffic. Each process is designed to be as simple as possible using custom web-based design tools.

The described modification of Hybrid code networks proved to be an efficient model for dialogue management by outperforming the state-of-the-art architecture on bAbI dialogue task 6. Thanks to the visualization of real user traffic within the designed tree structures of sub-dialogues, we are able to discover unexpected utterances and path inside the dialogues, and we can easily adapt existing models to handle them.

\bibliographystyle{iso690}
\bibliography{alquist}

\end{document}